\documentclass[11pt]{article}

\usepackage{microtype} 
\usepackage{booktabs}  

\usepackage{amsmath}
\usepackage{amsthm}
\usepackage[final]{automl}
\usepackage{xspace}

\usepackage{enumitem}

\usepackage{amsmath,amsfonts,bm}

\def\eqref#1{equation~\ref{#1}}

\def\1{\bm{1}}

\DeclareMathAlphabet{\mathsfit}{\encodingdefault}{\sfdefault}{m}{sl}
\SetMathAlphabet{\mathsfit}{bold}{\encodingdefault}{\sfdefault}{bx}{n}

\usepackage{natbib}

\usepackage{bigfoot}
\DeclareNewFootnote[para]{default}

\usepackage{hyperref}
\usepackage{url}

\usepackage{verbatim}
\usepackage{graphicx}
\usepackage{float}
\usepackage{cleveref}
\usepackage{wrapfig}
\usepackage{xcolor}
\usepackage{todonotes}

\usepackage[font=small,labelfont=bf]{caption}

\definecolor{codegreen}{rgb}{0,0.6,0}
\definecolor{codegray}{rgb}{0.5,0.5,0.5}
\definecolor{codepurple}{rgb}{0.58,0,0.82}
\definecolor{backcolour}{rgb}{0.95,0.95,0.92}
\usepackage{listings}
\lstdefinestyle{mystyle}{
    backgroundcolor=\color{backcolour},   
    commentstyle=\color{codegreen},
    keywordstyle=\color{magenta},
    numberstyle=\tiny\color{codegray},
    stringstyle=\color{codepurple},
    basicstyle=\ttfamily\footnotesize,
    breakatwhitespace=false,         
    breaklines=true,                 
    captionpos=b,                    
    keepspaces=true,                 
    numbers=left,                    
    numbersep=5pt,                  
    showspaces=false,                
    showstringspaces=false,
    showtabs=false,                  
    tabsize=2
}

\title{Open Source Vizier: Distributed Infrastructure and API for Reliable and Flexible Blackbox Optimization}

\author{Xingyou Song, Sagi Perel, Chansoo Lee, Greg Kochanski, Daniel Golovin \\
Google Research, Brain Team\\
}

\usepackage{color}
\definecolor{red}{rgb}{1,0,0}
\definecolor{darkred}{rgb}{0.5,0,0}
\definecolor{orange}{rgb}{1.0,0.64,0}
\definecolor{darkgreen}{rgb}{0,0.5,0}
\definecolor{darkblue}{rgb}{0,0,0.7}
\definecolor{purple}{rgb}{.6, 0,.6}
\newcount\Comments  %
\newcount\Edits  %
\newcommand{\edit}[1]{\ifnum\Edits=1\textcolor{blue}{#1}\else{#1}\fi}

\newcommand{\Study}{\texttt{Study}\xspace}
\newcommand{\Studies}{\texttt{Studies}\xspace}
\newcommand{\Trial}{\texttt{Trial}\xspace}
\newcommand{\Trials}{\texttt{Trials}\xspace}

\newcommand{\Suggestions}{\texttt{Suggestions}\xspace}
\newcommand{\Policy}{\texttt{Policy}\xspace}
\newcommand{\PolicySupporter}{\texttt{PolicySupporter}\xspace}

\begin{document}
\maketitle


\begin{abstract}
Vizier is the de-facto blackbox \edit{and hyperparameter} optimization service across Google, having optimized some of Google's largest products and research efforts. To operate at the scale of tuning thousands of users' critical systems, \edit{Google} Vizier solved key design challenges in providing multiple different features, while remaining fully fault-tolerant. In this paper, we introduce Open Source (OSS) Vizier, a \edit{standalone} Python-based interface for blackbox optimization and research, based on the Google-internal Vizier infrastructure and framework. OSS Vizier provides an API capable of defining and solving a wide variety of optimization problems, including multi-metric, early stopping, transfer learning, and conditional search. Furthermore, it is designed to be a distributed system that assures reliability, and allows multiple parallel evaluations of the user's objective function. The flexible RPC-based infrastructure allows users to access OSS Vizier from binaries written in any language. OSS Vizier also provides a back-end ("Pythia") API that gives algorithm authors a way to interface new algorithms with the core \edit{OSS} Vizier system. OSS Vizier is available at \url{https://github.com/google/vizier}.
\end{abstract}

\section{Introduction}

Blackbox optimization is the task of optimizing an objective function $f$ where the output $f(x)$ is the only available information about the objective. Due to its generality, blackbox optimization has been applied to an extremely broad range of applications, including but not limited to hyperparameter optimization \citep{he2021automl}, drug discovery \citep{bayesopt_chemisty}, reinforcement learning \citep{autorl_survey}, and industrial engineering \citep{materials_design}.

\begin{wrapfigure}[6]{r}{0.3\textwidth}
\vspace{-30pt}
\begin{center}
\includegraphics[keepaspectratio, width=0.2\textwidth]{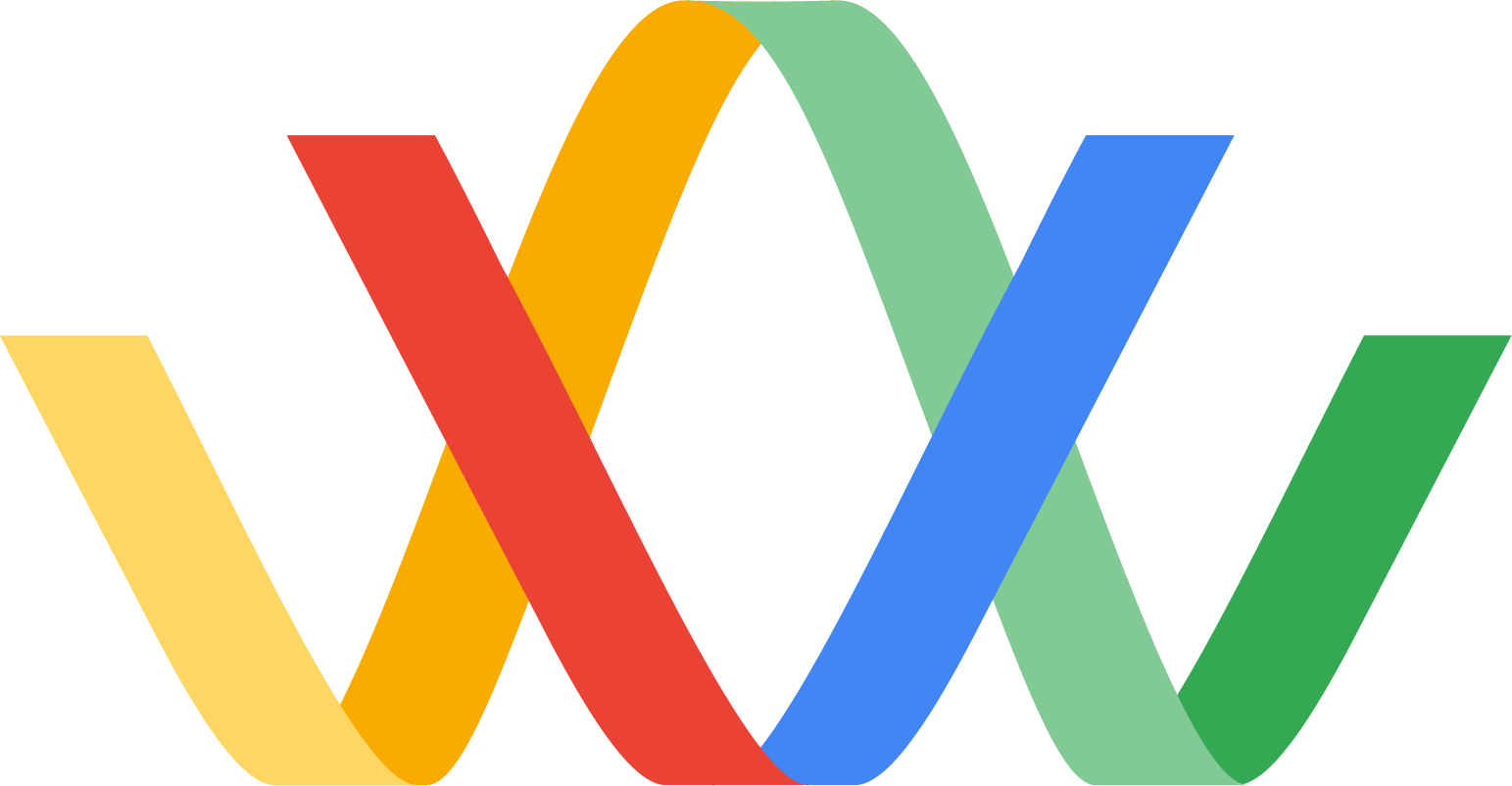}
\end{center}
\caption{OSS Vizier logo.}
\label{fig:vizier_logo} 
\end{wrapfigure}

Google Vizier \citep{vizier_v1} is the first hyperparameter tuning service designed to scale, and has thousands of monthly users both on the research\footnote{A list of research works that have used \edit{Google} Vizier can be found in Appendix \ref{sec:vizier_users}.} and production side at Google. Since its inception, Google Vizier has run millions of blackbox optimization tasks and saved a significant amount of computing and human resources to Google and its customers.

This paper describes Open Source (OSS) Vizier, a \edit{standalone} Python implementation of Google Vizier's APIs. It consists of a \textit{user API}, which allows users to configure and optimize their objective function, and a \textit{developer API}, which defines abstractions and utilities for implementing new optimization algorithms. Both APIs consist of Remote Procedure Call (RPC) protocols (Section \ref{sec:infra}) to allow the setup of a scalable, fault-tolerant and customizable blackbox optimization system, and Python libraries (Sections \ref{sec:pyvizier} and \ref{sec:pythia}) to abstract away the corresponding RPC protocols.

Compared to \citep{vizier_v1}, OSS Vizier features an evolved backend design for algorithm implementations, as well as new functionalities such as conditional search and multi-objective optimization. OSS Vizier's RPC API is based on Vertex Vizier\footnote{\url{https://cloud.google.com/vertex-ai/docs/vizier/overview}.}, making OSS Vizier compatible with any framework which integrates with Vertex Vizier, such as XManager\footnote{\url{https://github.com/deepmind/xmanager}.}.

\edit{Due to the existence of 3 different versions (Google, Vertex/Cloud, OSS) of Vizier, to prevent confusion, we explicitly refer to the version (e.g. "Google Vizier") whenever Vizier is mentioned. We summarize the distinct functionalities of each version of Vizier below:}

\begin{itemize}
\item \edit{Google Vizier: C++ based service hosted on Google’s internal servers and integrated deeply with Google's internal infrastructure. The service is available only for Google software engineers and researchers to tune their own objectives with a default algorithm.}
\item \edit{Vertex/Cloud Vizier: C++ based service hosted on Google Cloud servers, available for external customers + businesses to tune their own objectives with a default algorithm.}
\item \edit{OSS Vizier: Fully standalone and customizable code that allows researchers to host a Python-based service on their own servers, for any downstream users to tune their own objectives.}
\end{itemize}

\section{Problem and Our Contributions}
Blackbox optimization has a broad range of applications. Inside Google, these applications include: optimizing existing systems written in a wide variety of programming languages; tuning the hyperparameters of a large ML model using distributed parallel processes \citep{survey_on_distributed_ml}; optimizing a non-computational objective, which can be e.g. physical, chemical, biological, mechanical, or even human-evaluated \citep{vizier_cookie}. \edit{Generally, such objectives $f(x)$ we are interested in optimizing possess a moderate number (e.g. several hundred) of parameters for the input $x$, may produce noisy evaluation measurements, and may not be smooth or continuous.}

\edit{Furthermore, the} blackbox optimization workflow greatly varies depending on the application. The evaluation latency can be anywhere between seconds and weeks, while the budget for the number of evaluations, or \Trials, varies from tens to millions. Evaluations can be done asynchronously (e.g. ML model tuning) or in synchronous batches (e.g. wet lab experiments). Furthermore, evaluations may fail due to transient errors and should be retried, or may fail due to persistent errors (e.g. $f(x)$ cannot be measured) and should not be retried. \edit{One may also wish to stop the evaluation process early after observing intermediate measurements (e.g. from a ML model's learning curve) in order to save resources.}

To handle all of these scenarios, OSS Vizier is developed as a \textbf{service}. The service architecture does not make assumptions on how \Trials are evaluated, but rather simply specifies a stable API for obtaining \edit{suggestions $x_{1}, x_{2}, ...$ to evaluate and report results as \Trials}. Users have the freedom to determine when to request trials, how to evaluate trials, and when to report back results.

Another advantage of the service architecture is that it can collect data and metrics over time. \edit{Google Vizier} runs as a central service, \edit{ and we track} usage patterns to inform our research agenda, and our extensive database of runs serves as a valuable dataset for research into meta-learning and multitask transfer learning. This allows users to transparently benefit from the resulting improvements we make to the system.

\subsection{Comparisons to Related Work}
\label{subsec:comparisons}

Table \ref{ossoptimizationpackages} contains a non-comprehensive list of open-source packages for blackbox optimization, \edit{focusing on} hyperparameter tuning. Overall, OSS Vizier API is compatible with many of the features present in other \edit{hyperparameter tuning} open-source packages. We did not include commercial services for hyperparameter tuning such as Microsoft Azure, Amazon SageMaker, SigOpt and Vertex Vizier. For a comprehensive review \edit{of hyperparameter tuning tools,} see \citep{he2021automl}. \edit{There are many other blackbox optimization tools not mentioned in Table \ref{ossoptimizationpackages}, including iterated racing \citep{irace, irace_thesis_again}, as well as heuristics and automation of algorithm designs \citep{automatic_component_design, local_search_holgar}; see more comparisons and usages in \citep{lindauer2022smac3, feurer2015}.}

We divide the open-source packages into three categories:

\begin{itemize}
    \item \textbf{Services} host algorithms in a server. OSS Vizier, Advisor \citep{advisor} and OpenBox \citep{openbox}, which are modeled after Google Vizier \citep{vizier_v1}, belong to this category. Services are more flexible and scalable than frameworks, at the cost of engineering complexities.
    \item \textbf{Frameworks} execute the entire optimization, including both the suggestion algorithm and user evaluation code. Ax \citep{ax} and HpBandSter \citep{hpbandster} belong to this category. While frameworks are convenient, they often require knowledge on the system being optimized, such as how to manage resources and perform proper initialization and shutdown.
    \item \textbf{Libraries} implement blackbox optimization algorithms. HyperOpt \citep{bergstra2013making}, Emukit \citep{emukit2019}, and BoTorch \citep{botorch} belong to this category. Libraries offer the most freedom but lack scalability features such as error recovery and distributed/asynchronous trial evaluations. Instead, libaries are often used as algorithm implementations for frameworks or services (e.g. BoTorch in Ax).
\end{itemize}
 



\begin{table}[t]
\begin{center}
\scalebox{0.8}{
\begin{tabular}{p{0.15\textwidth}p{0.15\textwidth}p{0.13\textwidth}p{0.1\textwidth}p{0.56\textwidth}}
 \hline
 Name & Type & Client Languages & Parallel Trials & Features* \\ [0.5ex] 
 \hline\hline
 OSS Vizier & Service & Any & Yes & Multi-Objective, Early Stopping, Transfer Learning, Conditional Search \\ 
 \hline
 \edit{SMAC} & \edit{Framework} & \edit{Python} & \edit{Yes} & \edit{Multi-Objective, Multi-fidelity, Early Stopping, Conditional Search, Parameter Constraints} \\
 \hline
 Advisor & Service & Any & Yes & Early Stopping \\ 
 \hline
 OpenBox & Service & Any & Yes & Multi-Objective, Early Stopping, Transfer Learning, Parameter Constraints  \\
 \hline
 HpBandSter & Framework & Python & Yes & \edit{Early Stopping, Conditional Search, Parameter Constraints} \\ 
 \hline
 Ax + BoTorch & Framework & Python & Yes & Multi-Objective, Multi-fidelity, Early Stopping, Transfer Learning, Parameter and Outcome Constraints \\
 \hline
 HyperOpt & Library & Python & No &  Conditional Search \\ 
 \hline
 Emukit & Library & Python & No & Multi-Objective, Multi-fidelity, Outcome Constraints \\ 
 \hline
\end{tabular}
}
\caption{Open Source Optimization Packages. *OSS Vizier supports the API only.}\label{ossoptimizationpackages}
\end{center}
\vspace{-25pt}
\end{table}

One major architectural difference between OSS Vizier and other services is that OSS Vizier's algorithms may run in a separate service and communicate via RPCs with the API server, which performs database operations. With a distributed backend setup, OSS Vizier can serve algorithms written in different languages, scale up to thousands of concurrent users, and continuously process user requests without interruptions during a server maintenance or update. 

Furthermore, there are other minor differences between the services. While OSS Vizier and OpenBox support distinguishing workers via the workers' logical \edit{IDs} (Section~\ref{sec:userapi}), Advisor does not. In addition, OSS Vizier's Python clients possess more sophisticated functionalities than Advisor's, while OpenBox lacks a client implementation and requires users to implement client code using framework-provided worker wrappers. OSS Vizier also emphasizes algorithm development, by providing a developer API called \emph{Pythia} (Section~\ref{sec:pythia}) and utility libraries for state recovery. Other features of OSS Vizier include:
\begin{itemize}
\item OSS Vizier is one of the first open-source AutoML systems simultaneously compatible with a large-scale industry production service, Vertex Vizier, via our PyVizier library (Section~\ref{sec:pyvizier}).
\item The backend of OSS Vizier is based on the standard Google Protocol Buffer library, one of the most widely used RPC formats, which allows extensive customizability. In particular, the client (i.e. blackbox function to be tuned) can be written in any language and is not restricted to machine learning models in Python.
\item OSS Vizier is extensively integrated with numerous other Google packages, such as Deepmind XManager for experiment management (Section~\ref{sec:integrations}).
\end{itemize}

\section{Infrastructure}
\label{sec:infra}
We briefly conceptually define a \textit{Study} as all relevant data pertaining to an entire optimization loop, \edit{a \textit{Suggestion} as a suggested $x$, and a \textit{Trial} containing both $x$ and the objective $f(x)$. Note that in the code, we use \Trial as a container to store both $x$ and $f(x)$ and thus, a \Trial without $f(x)$ is also considered a suggestion.} We define these core primitives more programatically in Section \ref{sec:primitives}.

\subsection{Protocol Buffers}
\label{subsec:protobufs}
OSS Vizier's APIs are RPC interfaces that carry protocol buffers, or \textit{protobufs/protos}\footnote{\url{https://github.com/protocolbuffers/protobuf}}, to allow simple and efficient inter-machine communication. The protos are language- and platform- independent objects for serializing structured data, which make building external software layers and wrappers onto the system straightforward. In particular, the user can provide their own: 
\begin{itemize}
\item \textbf{Visualization Tools:} Since OSS Vizier securely stores all study data in its database, the data can then be loaded and visualized, with e.g. standard Python tools (Colab, Numpy, Scipy, Matplotlib) and other statistical packages such as R via RProtoBuf \citep{r_protobuf}. Front-end languages such as Angular/Javascript may also be used for visualizing studies. 
\item \textbf{Persistent Datastore:} The database in \edit{OSS} Vizier can changed based on the user's needs. For instance, a SQL-based datastore with full query functionality may be used to store study data.
\item \textbf{Clients:} Protobufs allow binaries written in Python, C++, and other languages to be tuned and/or used for evaluating the objective function. This allows \edit{OSS} Vizier to easily tune existing systems.
\end{itemize}
We explain the interactions between these components in a distributed backend below.

\subsection{Distributed Backend}
\label{sec:distributed-backend}
In order to serve multiple users while remaining fault-tolerant, \edit{OSS} Vizier runs in a distributed fashion, with a \textit{server} performing the algorithmic proposal work, while users or \textit{clients} communicate with the server via RPCs using the Client API, built upon gRPC \footnote{\url{https://grpc.io/}}. A packet of RPC communication is formatted in terms of standard Google protobufs.

\begin{figure}[h]
    \center
    \includegraphics[keepaspectratio, width=0.9\textwidth]{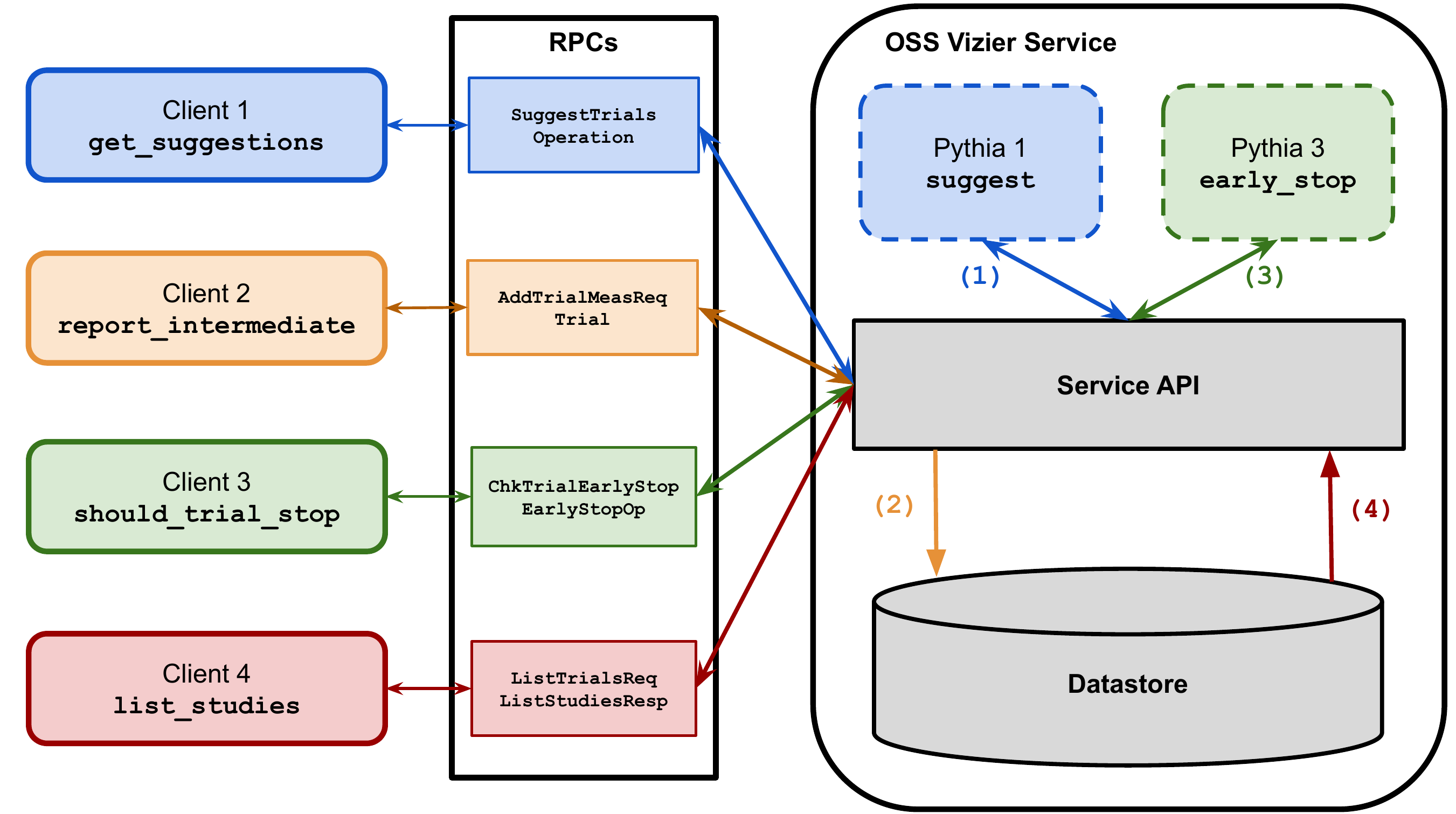}
\caption{Pictorial representation of the distributed pipeline. The OSS Vizier server services multiple clients, each with their own types of requests. Such requests can involve running Pythia Policies, saving measurement data, or retrieving previous studies. Note that Pythia may run as a separate service from the API service.}
\label{fig:vizier_infra}
\end{figure}

To start an optimization loop, a client will send a \texttt{CreateStudy} RPC request to the server, and the server will create a new Study in its datastore and return the ID to the client. The main tuning workflow in \edit{OSS} Vizier will then involve the following repeated cycle of events:

\begin{enumerate}
\item The client sends a \texttt{SuggestTrials} RPC request to the server.
\item The server creates a \texttt{Operation} in its datastore, and starts a thread to launch a Pythia policy (i.e. blackbox optimization algorithm) to compute the next suggested \Trials. The server returns an \texttt{Operation} protobuf to the client to denote the computation taking place.
\item The client will repeatedly poll the server via \texttt{GetOperation} RPCs to check the status of the \texttt{Operation} until the \texttt{Operation} is done.
\item When the Pythia policy produces its suggestions, the server will store these suggestions into the \texttt{Operation} and mark the \texttt{Operation} done, which will be collected by the client's \texttt{GetOperation} ping.
\item The client retrieves \edit{the suggestions} $x_{i}, ..., x_{i+n}$ stored inside the \texttt{Operation}, and returns objective function measurements $f(x_{i}),...,f(x_{i+n})$ to the server via calls to the \texttt{CompleteTrial} RPC.
\end{enumerate}

\edit{Note that the server may be launched in the same local process as the client, in cases where distributed computing is not needed and functio evaluation is cheap (e.g. benchmarking algorithms on synthetic functions). However, if the user wishes to use the distributed setting, the following are core advantages of OSS Vizier's system:}

\paragraph{Server-side Fault Tolerance} The \texttt{Operations} are stored in the database and contain sufficient information to restart the computation after a server crash, reboot, or update.

\paragraph{Automated/Early Stopping} A similar sequence of events takes place when the client sends a \texttt{CheckTrialEarlyStoppingStateRequest} RPC, in which the policy determines if a trial's evaluation should be stopped, and returns this signal as a boolean via the \texttt{EarlyStoppingOperation} RPC.

\paragraph{Batched/Parallel Evaluations} Note that \textit{multiple clients may work on the same study, and the same \edit\Trial.} This is important for compute-heavy experiments (e.g. neural architecture search) which need to parallelize workload by using multiple machines, with each machine $j$ evaluating the objective $f(x_{j})$ after being given suggestion $x_{j}$ from the server. 

\paragraph{Client-side Fault Tolerance} When one of the parallel workers fails and then reboots, \edit{the service} will assign the worker the same \edit{suggestion} as before. The worker can choose to load a model from the checkpoint to warm-start the evaluation.

\section{Core Primitives}
\label{sec:primitives}
\begin{figure}[h]
    \center
    \includegraphics[keepaspectratio, width=0.85\textwidth]{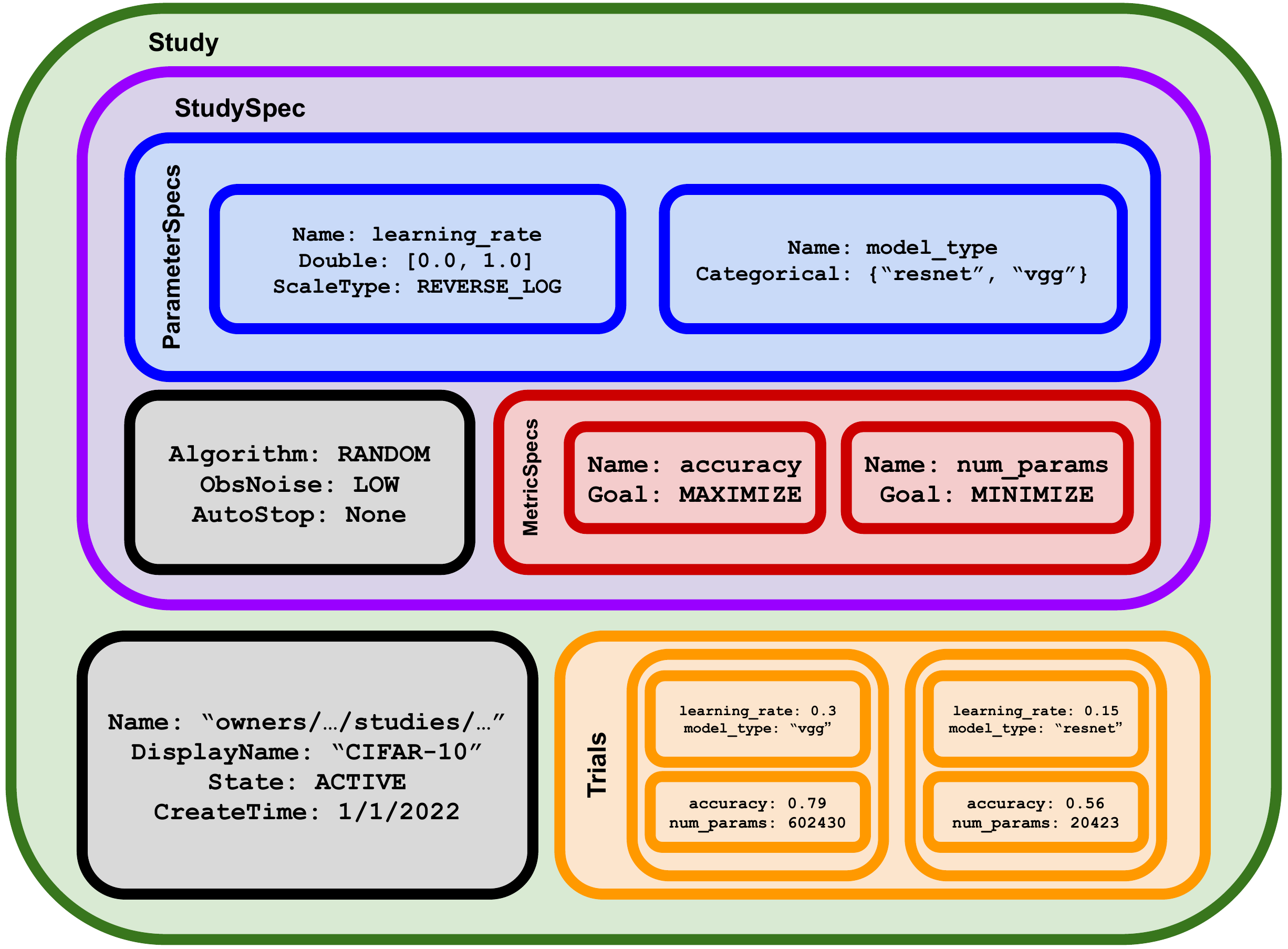}
\caption{Example of a study that tunes a deep learning task, featuring relevant data types.}
\label{fig:definitions}
\end{figure}

In Figure \ref{fig:definitions}, we provide a pictorial example representation of how OSS Vizier's primitives are structured; below we provide definitions.

\subsection{Definitions}
A \texttt{Study} is a single optimization run over a feasible space. Each study contains a name, its description, its state (e.g. \texttt{ACTIVE}, \texttt{INACTIVE}, or \texttt{COMPLETED}), a \texttt{StudySpec}, and a list of suggestions and evaluations (\texttt{Trials}).

A \texttt{StudySpec} contains the configuration details for the \texttt{Study}, namely the search space $\mathcal{X}$ (constructed by \texttt{ParameterSpecs}; see \S \ref{subsec:ss}), the algorithm to be used, automated stopping type (see Appendix \ref{subsec:automated_stopping}), the type of \texttt{ObservationNoise} (see Appendix \ref{subsec:obs_noise}), and at least one \texttt{MetricSpec}, containing information about the metric $f$ to optimize, including the metric name and the goal (i.e. whether to minimize or maximize $f$). Multiple \texttt{MetricSpecs} will be used for cases involving multiobjective optimization, where the goal is to find Pareto frontiers over multiple objectives $f_{1},...,f_{k}$.

A \texttt{Trial} is \edit{a container for the input $x \in \mathcal{X}$, as well as potentially the scalar value $f(x)$ or multiobjective values $f_{1}(x),...,f_{k}(x)$.} Each \edit{\Trial} possesses a \texttt{State}, which indicates what stage of the optimization process the \edit{\Trial} is in, with the two primary states being \texttt{ACTIVE} (meaning that $x$ has been suggested but not yet evaluated) and \texttt{COMPLETED} (meaning that evaluation is finished, and typically that the objectives $\left(f_{1}(x), ...,f_{k}(x)\right)$ have been calculated).  

Both the \texttt{StudySpec} and the \texttt{Trials} can contain \texttt{Metadata}. Metadata is not interpreted by \edit{OSS} Vizier, but rather \edit{a convenient method for developers} to store algorithm state, by users to store small amounts of arbitrary data, or as an extra communication medium between user code and algorithms.

\subsection{Search Space}
\label{subsec:ss}
Search spaces can be built by combining the the following primitives, or \texttt{ParameterSpec}s:

\begin{itemize}
\item \texttt{Double:} Specifies a continuous range of possible values in the closed interval $[a,b]$ for some real values $a \le b$.
\item \texttt{Integer:} Specifies an integer range of possible values in $[a,b] \subset \mathbb{Z}$ for some integers $a \le b$.
\item \texttt{Discrete:} Specifies a finite, ordered set of values from $\mathbb{R}$.
\item \texttt{Categorical:} Specifies an unordered list of strings.
\end{itemize}

Furthermore, each of the numerical parameters \texttt{\{Double, Integer, Discrete\}} has a \textit{scaling type}, which toggles whether the underlying algorithm is performing optimization in a transformed space. The scale type allows the user to conveniently inform the optimizer about the shape of the function, and can sometimes drastically accelerate the optimization. For instance, a user may use logarithmic scaling, which expresses the intent that a parameter ranging over $[0.001, 10]$ should roughly receive the same amount of attention in the subrange $[0.001, 0.01]$ as \edit{$[1, 10]$}, which would otherwise not be the case when using uniform scaling.

Each parameter also can potentially contain a list of child parameters, each of which will be active only if the parent's value matches the correct value(s). This allows the notion of \textit{conditional search}, which is helpful when dealing with search spaces involving incompatible parameters or parameters which only exist in specific scenarios.  For example, this can be useful when competitively tuning several machine learning algorithms along with each algorithm's parameters. E.g. one could tune the following for the \texttt{model} parameter: \texttt{\{"linear", "DNN", "random\_forest"\}}, each with its own set of parameters. Conditional parameters help keep the user's code organized, and also describe certain invariances to \edit{OSS} Vizier, namely that when \texttt{model="DNN"}, $f(x)$ will be independent of the \texttt{"random\_forest"} and \texttt{"linear"} model parameters.

These parameter primitives can be used flexibly to build highly complex search spaces, of which we provide examples in Appendix \ref{appendix:complex_ss}.

\subsection{PyVizier}
\label{sec:pyvizier}
All the above objects are implemented as protos to allow RPC exchanges through the service, as mentioned in Section \ref{sec:infra}. However, for ease-of-access, each object is also represented by an equivalent \textit{PyVizier} class to provide a more Pythonic interface, validation, and convenient construction (further details and examples are found in Appendix \ref{subsec:python_proto}). Translations to and from protos are provided by the \texttt{to\_proto()} and \texttt{from\_proto()} methods in PyVizier classes. \textbf{PyVizier provides a common interface across all Vizier variants (i.e. Google Vizier, Vertex Vizier, and OSS Vizier)}\footnote{For compatibility reasons, protos have slightly different names than PyVizier equivalents; e.g. \texttt{StudySpec} protos are equivalent to \texttt{StudyConfig} PyVizier objects. We describe conversions further in Appendix \ref{subsec:python_proto}}. The two intended primary use cases for PyVizier are: 
\begin{itemize}
\item Tuning user binaries. For such cases, the core PyVizier primitive is the \texttt{VizierClient} class that allows \edit{communication with the service.}
\item Developing algorithms for researchers. In this case, the core PyVizier primitives are the Pythia \texttt{Policy} and \texttt{PolicySupporter} classes.
\end{itemize}
Both cases typically use the \texttt{StudyConfig} and \texttt{SearchSpace} classes to define the optimization, and the \texttt{Trial}, and \texttt{Measurement} classes to support the evaluation.
We describe the two cases in detail below.

\section{User API: Parallel Distributed Tuning with \edit{OSS} Vizier Client}
\label{sec:userapi}
\begin{lstlisting}[label={lst:client}, float=h, language=Python, frame=tlrb, escapechar=|, captionpos=b, caption={Pseudocode for tuning a blackbox function using the included Python client. To save space, we did not use longer official argument names from the actual code.}]
from vizier import StudyConfig, VizierClient

config = StudyConfig() # Search space, metrics, and algorithm.
root = config.search_space.root # "Root" params must exist in every trial.
root.add_float('learning_rate', min=1e-4, max=1e-2, scale='LOG')
root.add_int('num_layers', min=1, max=5)
config.metrics.add('accuracy', goal='MAXIMIZE', min=0.0, max=1.0)
config.algorithm = 'RANDOM_SEARCH' 

client = VizierClient.load_or_create_study(
    'cifar10', config, client_id=sys.argv[1]) # Each client should use a unique id. |\label{lst:client:line:client_id_main}|
while suggestions := client.get_suggestions(count=1)
  # Evaluate the suggestion(s) and report the results to Vizier.
  for trial in suggestions:
    metrics = _evaluate_trial(trial.parameters)
    client.complete_trial(metrics, trial_id=trial.id)
\end{lstlisting}
The \edit{OSS} Vizier service must be set up first (see pseudocode in Appendix \ref{subsec:service_setup}), preferably on a \edit{multithreaded} machine capable of processing multiple RPCs \edit{concurrently}. Then, replicas of Code~Block~\ref{lst:client} can be launched in parallel, each with a unique command-line argument to be used as the client id in Line~\ref{lst:client:line:client_id_main}. The first replica to be launched creates a new \Study from the \texttt{StudyConfig}, which defines the search space, relevant metrics to be evaluated, and the algorithm for providing suggestions. The other replicas then load the same study to be worked on. There are a few important aspects worth noting in this setting:

\begin{itemize}
    \item \edit{The service} does not make any assumptions about how \Trials are evaluated. Users may complete \Trials at any latency, and may do so with a custom client written in any language. \edit{Algorithms may however, set a time limit and reassign \Trials to other clients to prevent stalling (e.g. due to a slow client).}
    \item Each \Trial is assigned a \texttt{client\_id} and only suggested to clients created with the same \texttt{client\_id}. This design makes it easy for users to recover from failures during \Trial evaluations; if one of the tuning binaries is accidentally shut down, users can simply restart the binary with the same client id. The tuning binary creates a new client attached to the same study and \edit{OSS} Vizier suggests the same \Trial.
    \item Multiple binaries can share the same \texttt{client\_id} and collaborate on evaluating the same \Trial. This feature is useful in tuning a large distributed model with multiple workers and evaluators.
    \item The client may optionally turn on automated stopping for objectives that can provide intermediate measurements (e.g. learning curves in deep learning applications). Further details and an example code snippet can be found in Appendix~\ref{subsec:automated_stopping} and Appendix~\ref{lst:autostop} respectively.
\end{itemize}

\section{Developer API: Implementing a New Algorithm Using Pythia Policy}
\label{sec:pythia}

\subsection{Overview}

As we have explained in Section~\ref{sec:infra}, OSS Vizier runs its algorithms in a binary called the \emph{Pythia service} (which can be the same binary as the API service). When the client asks for suggestions or early stopping decisions, the API service creates operations and sends requests to the Pythia service. This section describes the default python implementation of the Pythia service included in the open-source package. 

The Pythia service creates a \emph{\Policy} object that executes the algorithm and returns the response. \Policy is designed to be a minimal and general-purposed interface built on top of PyVizier, to allow researchers to quickly incorporate their own blackbox optimization algorithms. \Policy is usually \edit{given} a \emph{\PolicySupporter}, which is a mini-client specialized in reading and filtering \Trials. As shown in Code Block~\ref{lst:pythia1}, a typical \Policy loads \Trials via \PolicySupporter and processes the request at hand.

\begin{lstlisting}[label={lst:pythia1}, float=h, language=Python, frame=tlrb, captionpos=b, caption={Pseudocode for implementing a Gaussian Process Bandit.}]
from vizier.pythia import Policy, PolicySupporter, SuggestRequest, SuggestDecisions

class MyPolicy(Policy):
  def __init__(self, policy_supporter: PolicySupporter):
    self.policy_supporter = policy_supporter  # Used to obtain old trials.

  def suggest(self, request: SuggestRequest) -> SuggestDecisions:
    """Suggests trials to be evaluated."""
    Xs, y = _trials_to_np_arrays(self.policy_supporter.GetTrials(
        status='COMPLETED')) # Use COMPLETED trials only.
    model = _train_gp(Xs, y)
    return _optimize_ei(model, request.study_config.search_space)
\end{lstlisting}

\subsection{PolicySupporter}
\edit{The \PolicySupporter allows the \Policy to actively decide what \Trials from what \Studies are needed to generate the next batch of \Suggestions.  Policies can meta-learn from potentially any \Study in the database by calling the \texttt{GetStudyConfig} and \texttt{GetTrials} methods. Beyond that, the Policy can request only the \Trials it needs; e.g. for algorithms that only need to look at newly evaluated \Trials, this can reduce the database work by orders of magnitude relative to loading all the \Trials.}

\subsection{State Saving via Metadata}

\edit{The primary application of Google Vizier \citep{vizier_v1} was optimizing a blackbox function that is expensive to evaluate. Over time, as Google Vizier became widely adopted, there was an increasing number of applications where users wished to evaluate cheap functions over a very large number of \Trials. Popular methods for these applications include evolutionary methods and local search methods, such as NSGA-II \citep{nsga2}, Firefly \citep{firefly2010}, and Harmony Search  \citep{harmony_search} to name a few (For a survey on meta-heuristics, see \cite{metaheuristic_survey}).} 
 
\edit{A typical algorithm in this category} iteratively updates its population pool and generates mutations to be suggested, both of which take constant time with respect to the number of previous trials, \edit{as opposed to e.g. cubic time when using Gaussian Processes in a Bayesian Optimization loop. Since the lifespan of a \Policy object is equivalent to that of one suggestion or early stopping operation, the algorithm would need to fetch all \Trials in the \Study and reconstruct its state in $O(\text{number of previous trials})$ time. This leads to slow and difficult-to-maintain implementations.}

\PolicySupporter provides an easy-to-use API for developers to send algorithm states into \edit{the} database as \texttt{Metadata}. \texttt{Metadata} is a key-value mapping with namespaces that help prevent key collisions. There are two tables for metadata in the database: one attached to the \texttt{StudySpec} and another to each \texttt{Trial}. A \Policy can restore its last saved state from metadata, reflect the recently added \Trials, and process the operation at hand. We provide example code for this functionality in Appendix~\ref{sec:app:pythia_evo}


\section{Integrations} \label{sec:integrations}
OSS Vizier is also compatible with multiple other interfaces developed at Google as well. These include: 
\begin{itemize}
\item Vertex Vizier whose Protocol Buffer definitions are exactly the same\footnote{\url{https://cloud.google.com/vertex-ai/docs/reference/rest/v1beta1/StudySpec}.} as OSS Vizier's. This consistency also allows a wide variety of other packages (discussed below) pre-integrated with Vertex Vizier to be used with minimal changes.
\item Deepmind XManager experiments currently can be tuned by Vertex Vizier\footnote{\url{https://github.com/deepmind/xmanager/tree/main/xmanager/vizier}.} through \texttt{VizierWorker}. This worker can also be directly connected to an OSS Vizier server to allow custom policies to manage experiments.
\item OSS Vizier will also be the core backend for PyGlove \citep{pyglove}\footnote{\url{https://github.com/google/pyglove}}, which is a symbolic programming language for AutoML, in particular facilitating combinational and evolutionary optimization which are common in neural architecture search applications.
\end{itemize}


\section{Conclusion, Limitations and Broader Impact Statement}
\label{sec:conclusion_limitations_broader}
\paragraph{Conclusion} We discussed the motivations and benefits behind providing OSS Vizier as a service in comparison to other blackbox optimization libraries, and described how our gRPC-based distributed back-end infrastructure may be deployed as a fault-tolerant yet flexible system that is capable of supporting multiple clients and diverse use cases. We further outlined our client-server API for tuning, our algorithm development Pythia API, and integrations with other Google libraries.

\paragraph{Limitations} Due to proprietary and legal concerns, we are unable to open-source the default algorithms used \edit{in Google Vizier and Cloud Vizier}. Furthermore, this paper intentionally does not discuss algorithms or benchmarks, as the emphasis is on the systems aspect of AutoML. Algorithms may easily be added as policies to OSS Vizier's collection over time from contributors.

\edit{OSS Vizier also may not be suitable for all problems within the very broad scope of blackbox optimization. For instance, if evaluating $f(x)$ is very cheap and fast (e.g. miliseconds), then the OSS Vizier service itself may dominate the overall cost and speed. Furthermore, for problems requiring very large numbers of parameters (e.g. 100K+) and evaluations (e.g. 1M+), such as training a large neural network with gradientless methods \citep{ars, es_million_params}, OSS Vizier can also be inappropriate, as such cases can overload the datastore memory with redundant trials which do not need to be kept track of.}

\paragraph{Broader Impact} While there are a rich collection of sophisticated and effective AutoML algorithms published every year, broad adoption to practical use cases still remains low, as only 7\% of the ICLR 2020 and NeurIPS 2019 papers used a tuning method other than random or grid search \citep{people_use_random_search}. In comparison, Google Vizier is widely used among multiple researchers at Google, including for conference submissions. We hope that the release of OSS Vizier and its similar benefits may significantly improve the reach of AutoML techniques to users.

In terms of potential negative impacts, optimization as a service encourages central storage of data with the attendant risks and benefits. \edit{For example, currently through the Client API, a user may request all studies associated with another users, which may cause security and privacy concerns. This may be fixed by limiting user access to only their own studies in the service logic.} \edit{Furthermore, the host of the service currently has full access to all client data, which is another potential privacy concern. However, from our experience with Google Vizier, the most impactful applications for clients typically occur when parameters and measurements correspond to aggregate data (e.g. the learning rate of a ML algorithm, or e.g. the number of threads in a server) rather than data that describes individuals.} \edit{Furthermore}, data received by \edit{OSS} Vizier can be obscured to a degree \edit{to reduce unwanted exposure to the host.} \edit{Most notably, names (e.g. study name, parameter and metric names) can be encrypted, and (within limits) differential privacy \citep{differential_privacy_survey} approaches, especially for databases \citep{sql_differential_privacy}, can be applied to the parameters values and measurements.}



\section*{Acknowledgements}
The Vizier team consists of: Xingyou Song, Sagi Perel, Chansoo Lee, Greg Kochanski, Richard Zhang, Tzu-Kuo Huang, Setareh Ariafar, Lior Belenki, Daniel Golovin, and Adrian Reyes. 

We further thank Emily Fertig, Srinivas Vasudevan, Jacob Burnim, Brian Patton, Ben Lee, Christopher Suter for Tensorflow Probability integrations, Daiyi Peng for PyGlove integrations, Yingjie Miao for AutoRL integrations, Tom Hennigan, Pavel Sountsov, Richard Belleville, Bu Su Kim, Hao Li, and Yutian Chen for open source and infrastructure help, and George Dahl, Aleksandra Faust, and Zoubin Ghahramani for discussions.

\clearpage
\section{Reproducibility Checklist}

\begin{enumerate}
\item For all authors\dots
  \begin{enumerate}
  \item Do the main claims made in the abstract and introduction accurately
    reflect the paper's contributions and scope?
    \answerYes{We discussed the motivations for why OSS Vizier is designed as a service, and outlined in detail its distributed infrastructure. We further demonstrated (with pseudocode) the two main usages of OSS Vizier, which are to tune users' objects via client-side API, and develop algorithms via Pythia.}
  \item Did you describe the limitations of your work?
    \answerYes{See Section \ref{sec:conclusion_limitations_broader}.}
  \item Did you discuss any potential negative societal impacts of your work?
    \answerYes{See Section \ref{sec:conclusion_limitations_broader}.}
  \item Have you read the ethics review guidelines and ensured that your paper
    conforms to them?
    \answerYes{Our paper follows all of the ethics review guidelines.}
  \end{enumerate}
\item If you are including theoretical results\dots
  \begin{enumerate}
  \item Did you state the full set of assumptions of all theoretical results?
    \answerNA{This is a systems paper.}
  \item Did you include complete proofs of all theoretical results?
    \answerNA{This is a systems paper.}
  \end{enumerate}
\item If you ran experiments\dots
  \begin{enumerate}
  \item Did you include the code, data, and instructions needed to reproduce the
    main experimental results, including all requirements (e.g.,
    \texttt{requirements.txt} with explicit version), an instructive
    \texttt{README} with installation, and execution commands (either in the
    supplemental material or as a \textsc{url})?
    \answerYes{We have provided a README, installation instructions with a \texttt{requirements.txt}, numerous integration and unit tests along with PyTypes which demonstrate each code snippet's function.}
  \item Did you include the raw results of running the given instructions on the
    given code and data?
    \answerYes{Our unit-tests demonstrate the expected results of running all components of our code.}
  \item Did you include scripts and commands that can be used to generate the
    figures and tables in your paper based on the raw results of the code, data,
    and instructions given?
    \answerNA{This is a systems paper.}
  \item Did you ensure sufficient code quality such that your code can be safely
    executed and the code is properly documented?
    \answerYes{Our code follows all standard industry-wide coding practices at Google, which include extensive unit tests with continuous integration, PyType and PyLint enforcement for code cleanliness, and peer review during code submission.}
  \item Did you specify all the training details (e.g., data splits,
    pre-processing, search spaces, fixed hyperparameter settings, and how they
    were chosen)?
    \answerNA{This is a systems paper.}
  \item Did you ensure that you compared different methods (including your own)
    exactly on the same benchmarks, including the same datasets, search space,
    code for training and hyperparameters for that code?
    \answerNA{This is a systems paper.}
  \item Did you run ablation studies to assess the impact of different
    components of your approach?
    \answerNA{This is a systems paper.}
  \item Did you use the same evaluation protocol for the methods being compared?
    \answerNA{This is a systems paper.}
  \item Did you compare performance over time?
    \answerNA{This is a systems paper.}
  \item Did you perform multiple runs of your experiments and report random seeds?
    \answerNA{This is a systems paper.}
  \item Did you report error bars (e.g., with respect to the random seed after
    running experiments multiple times)?
    \answerNA{This is a systems paper.}
  \item Did you use tabular or surrogate benchmarks for in-depth evaluations?
    \answerNA{This is a systems paper.}
  \item Did you include the total amount of compute and the type of resources
    used (e.g., type of \textsc{gpu}s, internal cluster, or cloud provider)?
    \answerNA{This is a systems paper.}
  \item Did you report how you tuned hyperparameters, and what time and
    resources this required (if they were not automatically tuned by your AutoML
    method, e.g. in a \textsc{nas} approach; and also hyperparameters of your
    own method)?
    \answerNA{This is a systems paper.}
  \end{enumerate}
\item If you are using existing assets (e.g., code, data, models) or
  curating/releasing new assets\dots
  \begin{enumerate}
  \item If your work uses existing assets, did you cite the creators?
    \answerYes{Our work wraps around other Google libraries such as the Cloud Vizier SDK and Deepmind XManager, which we provided url links for.}
  \item Did you mention the license of the assets?
    \answerYes{Both the Cloud Vizier SDK and Deepmind XManager use the Apache 2.0 License.}
  \item Did you include any new assets either in the supplemental material or as
    a \textsc{url}?
    \answerNA{No new assets were used.}
  \item Did you discuss whether and how consent was obtained from people whose
    data you're using/curating?
    \answerNA{No human data was used.}
  \item Did you discuss whether the data you are using/curating contains
    personally identifiable information or offensive content?
    \answerNA{This is a systems paper without data use.}
  \end{enumerate}
\item If you used crowdsourcing or conducted research with human subjects\dots
  \begin{enumerate}
  \item Did you include the full text of instructions given to participants and
    screenshots, if applicable?
    \answerNA{Not applicable.}
  \item Did you describe any potential participant risks, with links to
    Institutional Review Board (\textsc{irb}) approvals, if applicable?
    \answerNA{Not applicable.}
  \item Did you include the estimated hourly wage paid to participants and the
    total amount spent on participant compensation?
    \answerNA{Not applicable.}
  \end{enumerate}
\end{enumerate}

\clearpage

\bibliography{references}
\bibliographystyle{apalike}

\newpage

\appendix
\section*{Appendix}

\section{Search Space Flexibility}
\label{appendix:complex_ss}
In this section, we describe the ways in which more complex search spaces may be created in OSS Vizier, showcasing its flexibility and applicability to a wide variety of problems.

\subsection{Combinatorial Optimization}
One of the most common uses for blackbox optimization in research involves combinatorial optimization. In this setting, $\mathcal{X}$ is usually defined via common manipulations over the set $[n] = \{0, 1,...,n-1\}$, such as permutations or subset selections. Below, we provide example methods to deal with such cases, in the order of most practical to least practical. We note that many of these methods are more suited for evolutionary algorithms which only need to utilize mutations and cross-overs between trials, rather than regression-based methods (e.g. Bayesian Optimization).

\subsubsection{Reparameterization}
Reparameterization of the search space $\mathcal{X}$ via conceptual means should be considered first, as it is one of the most practical and easiest ways to reduce the complexity of representing $\mathcal{X}$ in \edit{OSS} Vizier. Mathematically speaking, the high level idea is to construct a more practical search space $\mathcal{Z}$ which can easily be represented in \edit{OSS} Vizier, and then create a surjective mapping $\Phi: \mathcal{Z} \rightarrow \mathcal{X}$.

For basic combinatorial objects such as permutations, if we consider the standard permutation space $\mathcal{X} = \{x : x \in [n]^{n}, x_{i} \neq x_{j} \> \forall i \neq j \}$, then we may define $\mathcal{Z} = [n] \times [n-1] \times ... \times [2] \times [1]$ and allow $\Phi$ to be the decoding operator for the Lehmer code\footnote{\url{https://en.wikipedia.org/wiki/Lehmer_code}}. If $\mathcal{X} = \{x: x \subseteq [n], |x| = k\}$ involves subset selection, then we may define $\mathcal{Z} = [n] \times [n-1] \times ... \times [n - k + 1]$ and apply a similar mapping.

Another common case involves searching over the space of graphs. In such scenarios, there are a multitude of methods to parameterizing the graph, including adjacency matrices via $[n] \times [n]$. An illustrative example can be seen across neural architecture search (NAS) benchmarks. Even though such search spaces correspond to graph objects, ironically, many NAS benchmarks, termed ``NASBENCH"s, actually do not use nested or conditional search spaces. For instance, NASBENCH-101 \citep{nasbench_101} uses only a flat adjacency matrix and flat operation list. NASBENCH-201 \citep{nasbench_201} is even simpler, as it takes the graph dual of the node-op representation, allowing the search space to be a full feasible set represented by only 5 categorical parameters.

\subsubsection{Infeasibility}
In some scenarios, we may not be able to find a mapping $\Phi$ as in the reparameterization case above, but instead may lift the search space $\mathcal{X}$ into a larger search space $\mathcal{Z}$, where $\mathcal{X} \subset \mathcal{Z}$, and thus perform search on $\mathcal{Z}$ instead. For trials in $\mathcal{Z} - \mathcal{X} = \{ z: z \in \mathcal{Z}, z \notin \mathcal{X} \}$, \edit{OSS} Vizier supports reporting these trials as infeasible. 
As a basic example, if $\mathcal{X} = \{ x \in \mathbb{R}^{2} : ||x|| \le 1\} $ defines a disk, then $\mathcal{Z} = [-1, 1]^{2}$. Another example can be seen with the same NASBENCH-101 \citep{nasbench_101} benchmark described earlier, where some pairs of adjacency matrices and operation lists do not correspond to an actual valid graph, and are thus infeasible. 

The main limitation is if $|\mathcal{X}| \ll |\mathcal{Z}|$, the vast bulk of trials may be infeasible, and if so, the search will converge slowly. Furthermore, for the disk case, this can lead to problems during optimization, as it creates a sharp border $\mathcal{X} \cap \mathcal{Z}$ and a flat infeasible region $\mathcal{Z} - \mathcal{X}$. This leads to lack of information about which infeasible points are better/worse than others, and can also make it difficult to find a small feasible region. Modelling techniques such Gaussian Processes also inherently assume the objective function is continuous everywhere, which is incompatible with the discontinuity from the border $\mathcal{X} \cap \mathcal{Z}$.

\subsubsection{Serialization}
If all else fails, we may avoid the use of the \textit{ParameterSpec} API and simply serialize $x \in \mathcal{X}$ into a string format, which can then be inserted into a Trial's \textit{metadata} field. In cooperation with a custom Pythia policy, this can be very effective.

\section{Additional \edit{OSS} Vizier Settings}


\subsection{Automated Stopping}
\label{subsec:automated_stopping}
Automated/early stopping is used commonly when trials can be stopped early to save resources, and is determined by the trial's intermediate measurements. Currently there are two modes to automated stopping which the client can specify in their \texttt{StudyConfig}:
\begin{itemize}
\item Decay Curve Automated Stopping, in which a Gaussian Process Regressor is built to predict the final objective value of a Trial based on the already completed Trials and the intermediate measurements of the current Trial. Early stopping is requested for the current Trial if there is very low probability to exceed the optimal value found so far.
\item Median Automated Stopping, in which a pending trial is stopped if the Trial's best objective value is strictly below the median 'performance' of all completed Trials reported up to the Trial's last measurement. Currently, 'performance' refers to the running average of the objective values reported by the Trial in each measurement.
\end{itemize}

\subsection{Observation Noise}
\label{subsec:obs_noise}
We have found it useful to let the user give Vizer a hint about the amount of noise in their evaluations via the \texttt{StudyConfig}.  Because the noise/irreproducibility of evaluations is often not well known in advance by users, we give users a broad choice that the noise is either \texttt{Low} or \texttt{High}:

\begin{itemize}
\item \texttt{Low}: This implies that the objective function is (nearly) perfectly reproducible, and an algorithm should never repeat the same Trial parameters.
\item \texttt{High}: This assumes there is enough noise in the evaluations that it is worthwhile for \edit{OSS} Vizier sometimes to re-evaluate with the same (or nearly) parameter values.
\end{itemize}

This hint is passed to the Pythia policy, and the policy is free to also use this hint to e.g. adjust priors on the hyperparameters of a Gaussian Process regressor.


\clearpage

\section{\edit{Google} Vizier Users and Citations}
\label{sec:vizier_users}
Besides Google Vizier's extensive internal production usage, below comprises a selected list of publicly available research works\footnote{Full list of \edit{Google} Vizier's citations: \url{https://scholar.google.com/scholar?oi=bibs&hl=en&cites=14342343058535677299}.} which have used Google Vizier, demonstrating its rich research user-base \edit{which may directly translate to OSS Vizier's future user-base as well.}

\paragraph{Neural Architecture Search} \edit{Google} Vizier has acted as a core backend for many of the neural architecture search (NAS) efforts at Google, beginning with \edit{Google} Vizier having been used to hyperparameter tune the RNN controller in the original NAS paper \citep{original_nas}. Over the course of NAS research, \edit{Google} Vizier has also been used to reliably handle the training of thousands of models \citep{barett_cvpr_nas, multiscale}, as well as comparisons against different NAS optimization algorithms in NASBENCH-101 \citep{nasbench_101}. Furthermore, it serves as the primary distributed backend for PyGlove \citep{pyglove}, a core evolutionary algorithm API for NAS research across Google.

\paragraph{Hardware and Systems} \edit{Google} Vizier's tuning led to crucial gains for hardware benchmarking, such as improving JAX's MLPerf scores over TPUs \footnote{Link too long; hyperlink can be found \href{https://cloud.google.com/blog/products/ai-machine-learning/google-breaks-ai-performance-records-in-mlperf-with-worlds-fastest-training-supercomputer}{\edit{here.}}}. \edit{Google} Vizier's multiobjective optimization capabilities were a key component in producing better computer architecture designs in APOLLO \citep{apollo} \footnote{\url{https://ai.googleblog.com/2021/02/machine-learning-for-computer.html}}. Furthermore, \edit{Google} Vizier was a key component to \textit{Full-stack Accelerator Search Technique} (FAST) \citep{accelerator_search}, an automated framework for jointly optimizing hardware datapath, software schedule, and compiler passes.

\paragraph{Reinforcement Learning} ``AutoRL" \citep{autorl_survey} has recently seen a great deal of promise in automating reinforcement learning systems. \edit{Google} Vizier was extensively used as the core component in tuning hyperparameters and rewards in navigation \citep{sandra_evolving_rewards_autorl, sandra_indoor_autorl, sandra_navigation_autorl}. \edit{Google} Vizier's backend was also used to host the Regularized Evolution optimizer \citep{regularized_evolution}, used for evolving RL algorithms \citep{evolving_rl_algorithms}, where the search space involved combinatorial directed acyclic graphs (DAGs). On the infrastructure side, \edit{Google} Vizier was used to improve the performance of Reverb \citep{reverb}, one of the core replay buffer APIs used for most RL projects at Google. \citep{contrastive}

\paragraph{Biology/Chemistry/Healthcare} \edit{Google} Vizier's algorithms were used for comparison on several papers related to protein optimization \citep{blundell}, and was also used to tune RNNs for peptide identification in \citep{peptide}. For healthcare, \edit{Google} Vizier was used to tune models for classifying diseases such as diabetic retinopathy \citep{diabetic_retinopathy}

\paragraph{General Deep Learning} For fundamental research, \edit{Google} Vizier was used to tune Neural Additive Models \citep{neural_additive}, and has also been the backbone of core research into infinite-width deep neural networks, having tuned \citep{roman_ntk_conv, roman_ntk_finite, roman_infinite_attention, roman_ntk_bnn}. For NLP-based tasks, \edit{Google} Vizier regularly tunes language model training, and has also been used to search feature weights in \citep{curriculum_NMT}, as well improve performance for work on theorem proving \citep{theorem_proving}. Computer vision models such as ones used for the Pixel-3\footnote{\url{https://ai.googleblog.com/2018/12/top-shot-on-pixel-3.html}} have been tuned by \edit{Google} Vizier.

\paragraph{Miscallaneous:} As an example of tuning for human-based judgement on objectives unrelated to technology, \edit{Google} Vizier was used to tune the recipe for cookie-baking \citep{vizier_cookie}.

\clearpage

\section{Extended Code Samples}

\subsection{Automated stopping}
Code Block \ref{lst:autostop} demonstrates the use of automated stopping, when training a standard machine learning model.

\begin{lstlisting}[label={lst:autostop}, float=h, language=Python, frame=tlrb, escapechar=|, captionpos=b, caption={\emph{Pseudocode} for tuning a model using the included Python client, with early stopping enabled.}]
from vizier import StudyConfig, VizierClient

config = StudyConfig()
... # configure search space and metrics
client = VizierClient.load_or_create_study(
    'cifar10', config, client_id=sys.argv[1]) # Each client should use a unique id. |\label{lst:client:line:client_id_appendix}|
while suggestions := client.get_suggestions(count=1)
  # Evaluate the suggestion(s) and report the results to OSS Vizier.
  for trial in suggestions:
    for epoch in range(EPOCHS):
      if client.should_trial_stop(trial.id):
         break
       metrics = model.train_and_evaluate(trial.parameters['learning_rate'])
       client.report_metrics(epoch, metrics)
    metrics = model.evaluate()
    client.complete_trial(metrics, trial_id=trial.id)
\end{lstlisting}

\subsection{Service Setup}
\label{subsec:service_setup}

Code Block \ref{lst:server} displays the simple method in which to setup the service on a multithreaded server.

\begin{lstlisting}[label={lst:server}, float=h, language=Python, frame=tlrb, captionpos=b, caption={Pseudocode for setting up the service on a server.}]
from vizier.service import vizier_server
from vizier.service import vizier_service_pb2_grpc

hostname = 'localhost' # Example; usually user-specified
port = 6006 # Example; usually user-specified
address = f'{hostname}:{port}' 
servicer = vizier_server.VizierService()

server = grpc.server(futures.ThreadPoolExecutor(max_workers=100))
vizier_service_pb2_grpc.add_VizierServiceServicer_to_server(servicer, server)
server.add_secure_port(address, grpc.local_server_credentials())
server.start()
\end{lstlisting}

\clearpage

\subsection{Proto vs Python API}
\label{subsec:python_proto}
We provide an example of equivalent methods between PyVizier and corresponding Protocol Buffer objects. Note that clients and algorithm developers should not normally need to modify protos. Such cases are more common if one wishes to add extra layers on top of the service, as mentioned in Subsection \ref{subsec:protobufs}.

\begin{lstlisting}[label={lst:proto_trial}, language=Python, float=h, language=Python, frame=tlrb, captionpos=b, caption={Original Protocol Buffer method of creating a Trial.}]
from vizier.service import study_pb2
from google.protobuf import struct_pb2

param_1 = study_pb2.Trial.Parameter(parameter_id='learning_rate', value=struct_pb2.Value(number_value=0.4))
param_2 = study_pb2.Trial.Parameter(parameter_id='model_type', value=struct_pb2.Value(string_value='vgg'))
metric_1 = study_pb2.Measurement.Metric(metric_id='accuracy',value=0.4)
metric_2 = study_pb2.Measurement.Metric(metric_id='num_params',value=20423)
final_measurement = study_pb2.Trial.Measurement(metrics=[metric_1,metric_2])
trial = study_pb2.Trial(parameters=[param_1,param_2], final_measurement=final_measurement)
\end{lstlisting}

\begin{lstlisting}[label={lst:pyvizier_trial}, language=Python, float=h, language=Python, frame=tlrb, captionpos=b, caption={Equivalent method of writing the PyVizier version of the Trial from Code Block \ref{lst:proto_trial}. Note the significantly more "Pythonic" way of writing code, with a significant reduction in code complexity.}]
from vizier.pyvizier import ParameterDict, ParameterValue, Measurement, Metric, Trial

params=ParameterDict()
params['learning_rate'] = ParameterValue(0.4)
params['model_type'] = ParameterValue('vgg')
final_measurement = Measurement()
final_measurement.metrics['accuracy'] = Metric(0.7)
final_measurement.metrics['num_params'] = Metric(20423)
trial = pv.Trial(parameters=params,final_measurement=final_measurement)
\end{lstlisting}

We also provide in Table \ref{table:pyvizier_names}, changes between \edit{OSS Vizier's Protocol Buffer names and their corresponding PyVizier names, as well as converter objects.} 

\begin{table}[h]
\centering
\begin{tabular}{ |c|c|c| } 
\hline
 Protocol Buffer Name & PyVizier Name & Converter \\
 \hline
 \hline
 \texttt{Study} & \texttt{Study} & N/A \\
 \texttt{StudySpec} & \texttt{SearchSpace} + \texttt{StudyConfig} & \texttt{SearchSpace} (self) + \texttt{StudyConfig} (self) \\ 
 ParameterSpec & \texttt{ParameterConfig} & \texttt{ParameterConfigConverter} \\
 \texttt{Trial} & \texttt{Trial} & \texttt{TrialConverter} \\
 \texttt{Parameter} & \texttt{ParameterValue} & \texttt{ParameterValueConvereter} \\ 
 \texttt{MetricSpec} & \texttt{MetricInformation} & \texttt{MetricInformation} (self) \\ 
 \texttt{Measurement} & \texttt{Measurement} & \texttt{MeasurementConverter} \\
 \hline
\end{tabular}
\caption{Corresponding names and conversion objects between Protocol Buffer and PyVizier objects. (self) denotes that the PyVizier object has its own immediate \texttt{to\_proto()} and \texttt{from\_proto()} functions.}
\label{table:pyvizier_names}
\end{table}

\clearpage

\subsection{Implementing an Evolutionary Algorithm}
\label{sec:app:pythia_evo}

OSS Vizier possesses an abstraction \texttt{SerializableDesigner} defined purely in terms of PyVizier without any Pythia dependencies. This interface wraps around most commonly used algorithms which sequentially update their internal states as new observations arrive. The interface is easy to understand and can be wrapped into a Pythia policy using the \texttt{SerializableDesignerPolicy} class which handles state management. See Code Block~\ref{lst:es_impl} for an example.

\begin{lstlisting}[label={lst:es_impl}, language=Python, float=h, language=Python, frame=tlrb, captionpos=b, caption={Example Pseudocode of implementing an evolutionary algorithm as a Pythia policy using \texttt{SerializableDesigner} interface.}]
from vizier import pyvizier as vz

class RegEvo(SerializableDesigner):

  # override
  def suggest(self, count: Optional[int]) -> Sequence[vz.TrialSuggestion]
    """Generate `count` number of mutations and return them."""
  
  # override
  def update(self, delta: CompletedTrials):
    """Apply selection step and update the population pool."""

  # override
  def dump(self) -> vz.Metadata:
    """Dumps the population pool."""
    md = vz.Metadata()
    md['population'] = json.dumps(...)
    return md
    
  # override
  def recover(cls: Type['_S'], metadata: vz.Metadata) -> '_S':
    """Restores the population pool."""
    if 'population' not in md:
      raise HarmlessDecodeError('Cannot find key: "population"')
    ... = json.loads(md['population'])
    
policy = SerializableDesignerPolicy(
    policy_supporter,
    designer_factory=RegEvo.__init__,
    designer_cls=RegEvo)
\end{lstlisting}

\end{document}